\pdfoutput=1
\documentclass[10pt,twocolumn,letterpaper]{article}

\usepackage{cvpr}
\usepackage{times}
\usepackage{epsfig}
\usepackage{graphicx}
\usepackage{amsmath}
\usepackage{amssymb}

\usepackage[pagebackref=true,breaklinks=true,letterpaper=true,colorlinks,bookmarks=false,citecolor=blue,linkcolor=green]{hyperref}

\cvprfinalcopy 

\def\cvprPaperID{****} 

\begin{document}

\title{Measuring Machine Intelligence Through Visual Question Answering}
\author{C. Lawrence Zitnick\\
\normalsize Facebook AI Research\\
{\tt\small zitnick@fb.com}
\and
Aishwarya Agrawal\\
\normalsize Virginia Tech\\
{\tt\small aish@vt.edu}
\and
Stanislaw Antol\\
\normalsize Virginia Tech\\
{\tt\small santol@vt.edu}
\and
Margaret Mitchell\\
\normalsize Microsoft Research\\
{\tt\small memitc@microsoft.com}
\and
Dhruv Batra\\
\normalsize Virginia Tech\\
{\tt\small dbatra@vt.edu}
\and
Devi Parikh\\
\normalsize Virginia Tech\\
{\tt\small parikh@vt.edu}
}

\maketitle

\begin{abstract}
As machines have become more intelligent, there has been a renewed interest in methods for measuring their intelligence. A common approach is to propose tasks for which a human excels, but one which machines find difficult. However, an ideal task should also be easy to evaluate and not be easily gameable. We begin with a case study exploring the recently popular task of image captioning and its limitations as a task for measuring machine intelligence. An alternative and more promising task is Visual Question Answering that tests a machine's ability to reason about language and vision. We describe a dataset unprecedented in size created for the task that contains over 760,000 human generated questions about images. Using around 10 million human generated answers, machines may be easily evaluated.
\end{abstract}

\section{Introduction}
Humans have an amazing ability to both understand and reason about our world through a variety of senses or modalities. A sentence such as ``Mary quickly ran away from the growling bear.'', conjures both vivid visual and auditory interpretations. We picture Mary running in the opposite direction of a ferocious bear with the sound of the bear being enough to frighten anyone. While interpreting a sentence such as this is effortless to a human, designing intelligent machines with the same deep understanding is anything but. How would a machine know Mary is frightened? What is likely to happen to Mary if she doesn't run? Even simple implications of the sentence, such as ``Mary is likely outside'' may be nontrivial to deduce.

How can we determine if a machine has achieved the same deep understanding of our world as a human? In our example sentence above, a human's understanding is rooted in multiple modalities. They can visualize a scene depicting Mary running, they can imagine the sound of the bear, and even how the bear's fur might feel when touched. Conversely, if shown a picture or even an auditory recording of a woman running from a bear, a human may similarly describe the scene. Perhaps machine intelligence could be tested in a similar manner? Can a machine use natural language to describe a picture similar to a human? Similarly, could a machine generate a scene given a written description? In fact these tasks have been a goal of artificial intelligence research since its inception. Marvin Minsky famously stated in 1966 \cite{Crevier1993} to one of his students,``Connect a television camera to a computer and get the machine to describe what it sees.''  At the time, and even today, the full complexities of this task are still being discovered.

\section{Image Captioning}

\begin{figure}
\centering
\includegraphics[width=1\linewidth]{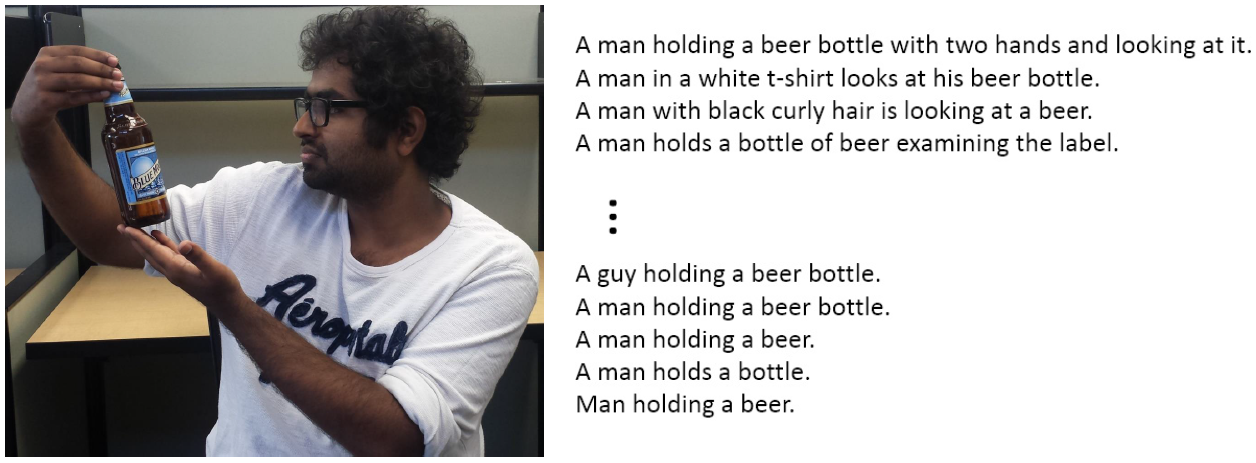}
\caption{Example image captions written for an image sorted by caption length.}
 \label{fig:length}
 \end{figure}

Are tasks such as image captioning \cite{barnard2001learning,kulkarni2011baby,mitchell2012midge,farhadi2010every,hodosh2013framing,captioning_msr,captioning_xinlei,captioning_berkeley,captioning_baidu_ucla,captioning_toronto,captioning_stanford,captioning_google} promising candidates for testing artificial intelligence? These tasks have advantages, such as being easy to describe and being capable of capturing the imagination of the public \cite{nyt}. Unfortunately, tasks such as image captioning have proven problematic as actual tests of intelligence. Most notably, the evaluation of image captions may be as difficult as the image captioning task itself \cite{elliott2014comparing,cider,hodosh2013framing,kulkarni2011baby,mitchell2012midge}. It has been observed that captions judged as ``good'' by human observers may actually contain significant variance even though they describe the same image \cite{cider}. For instance see Figure \ref{fig:length}. Many people would judge the longer more detailed captions as better. However, the details described by the captions varies significantly, e.g. ``two hands'', ``white t-shirt'', ``black curly hair'', ``label'', etc.  How can we evaluate a caption if there is no consensus on what should be contained in a ``good'' caption? However, for shorter less detailed captions that are commonly written by humans a rough consensus is achieved ``A man holding a beer bottle.'' This leads to the somewhat counterintuitive conclusion that captions humans like aren't necessarily ``human-like''.

\begin{figure}
\centering
\includegraphics[width=1\linewidth]{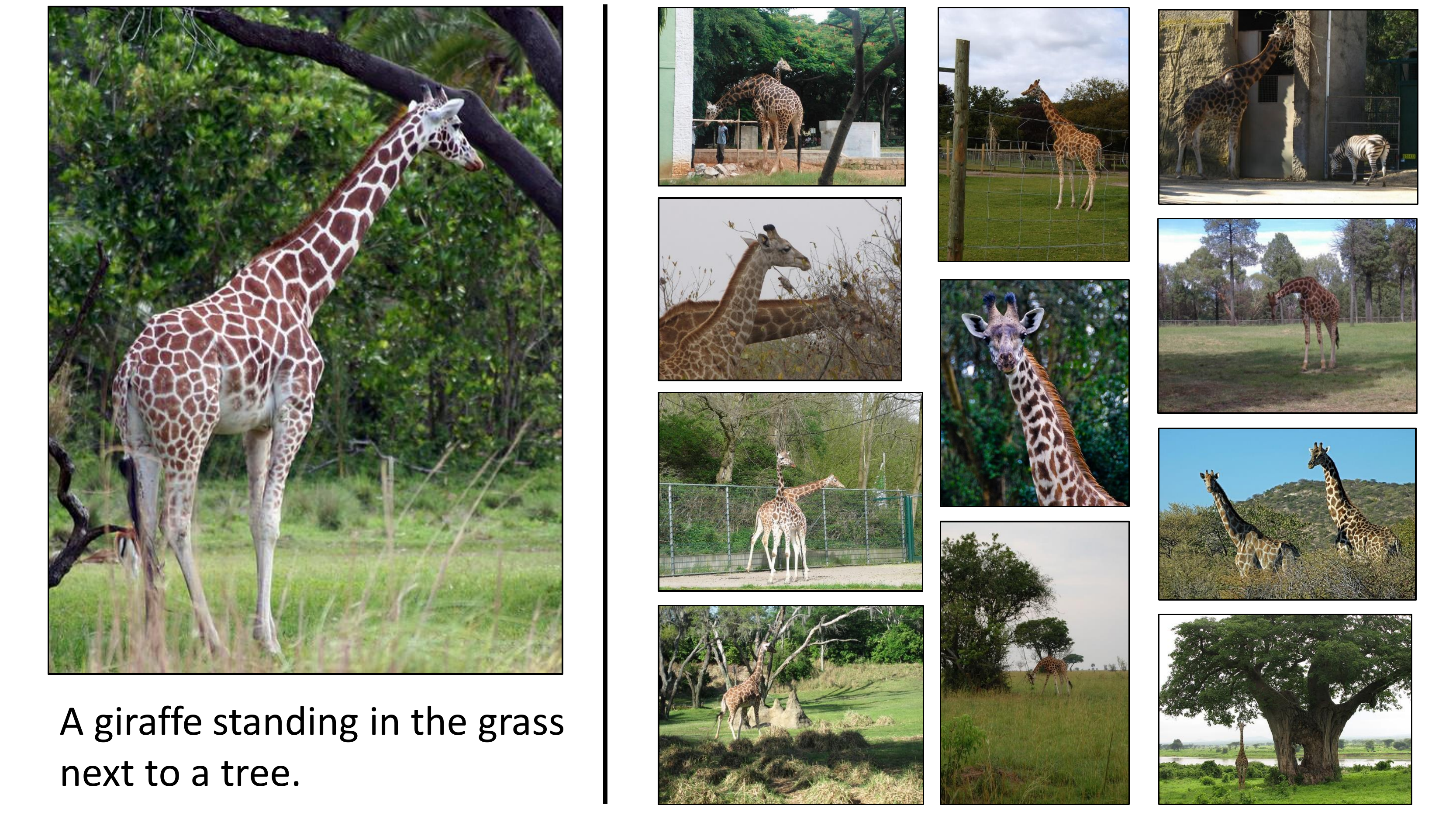}
\caption{(left) An example image caption generated from \cite{captioning_msr}. (right) A set of semantically similar images in the MS COCO training dataset for which the same caption could apply.}
 \label{fig:msr}
 \end{figure}

The task of image captioning also suffers from another less obvious drawback. In many cases it might be too easy! Consider an example success from a recent paper on image captioning \cite{captioning_msr}, Figure \ref{fig:msr}. Upon first inspection this caption appears to have been generated from a deep understanding of the image. For instance, in Figure \ref{fig:msr} the machine must have detected a giraffe, grass and tree. It understood that the giraffe was standing, and the thing it was standing on was grass. It knows the tree and giraffe are ``next to'' each other, etc. Is this interpretation of the machine's depth of understanding correct?
When judging the results of an AI system, it is not only important to analyze its output, but the data used for its training. The results in Figure \ref{fig:msr} were obtained by training on the Microsoft Common Objects in Context (MS COCO) dataset \cite{coco}. This dataset contains five independent captions written by humans for over 120,000 images \cite{COCOCaptions}. If we examine the image in Figure \ref{fig:msr} and the images in the training dataset we can make an interesting observation. For many testing images, there exists a significant number of semantically similar training images, Figure \ref{fig:msr}(right). If two images share enough semantic similarity, it is possible a single caption could describe them both.

This observation leads to a surprisingly simple algorithm for generating captions \cite{kNN}. Given a test image, collect a set of captions from images that are visually similar. From this set, select the caption with highest consensus \cite{cider}, i.e. the caption most similar to the other captions in the set. In many cases the consensus caption is indeed a good caption. When judged by humans, $21.6\%$ of these borrowed captions are judged to be equal to or better than those written by humans for the image specifically. Despite its simplicity, this approach is competitive with more advance approaches using recurrent neural networks \cite{captioning_xinlei,captioning_berkeley,captioning_baidu_ucla,captioning_toronto,captioning_stanford,captioning_google} and other language models \cite{captioning_msr} which can achieve $27.3\%$ when compared to human captions. Even methods using recurrent neural networks commonly produce captions that are identical to training captions even though they're not explicitly trained to do so. If captions are ``generated'' by borrowing them from other images, these algorithms are clearly not demonstrating a deep understanding of language, semantics and their visual interpretation. The odds of two humans repeating a sentence is quite rare.

One could make the case that the fault is not with the algorithms but in the data used for training. That is, the dataset contains too many semantically similar images. However, even in randomly sampled images from the web, a photographer bias is found. Humans capture similar images to each other. Many of our tastes or preferences are universal.

\section{Visual Question Answering}
 
As we demonstrated using the task of image captioning, determining a multimodal task for measuring a machine's intelligence is challenging. The task must be easy to evaluate, yet hard to solve. That is, it's evaluation shouldn't be as hard as the task itself, and it must not be solvable using ``shortcuts'' or ``cheats''. To solve these two problems we propose the task of Visual Question Answering (VQA) \cite{VQA,geman,fritz,SongChun_video_queries,vizwiz,GaoMZHWX15}.

\begin{figure}
\centering
\includegraphics[width=1\linewidth]{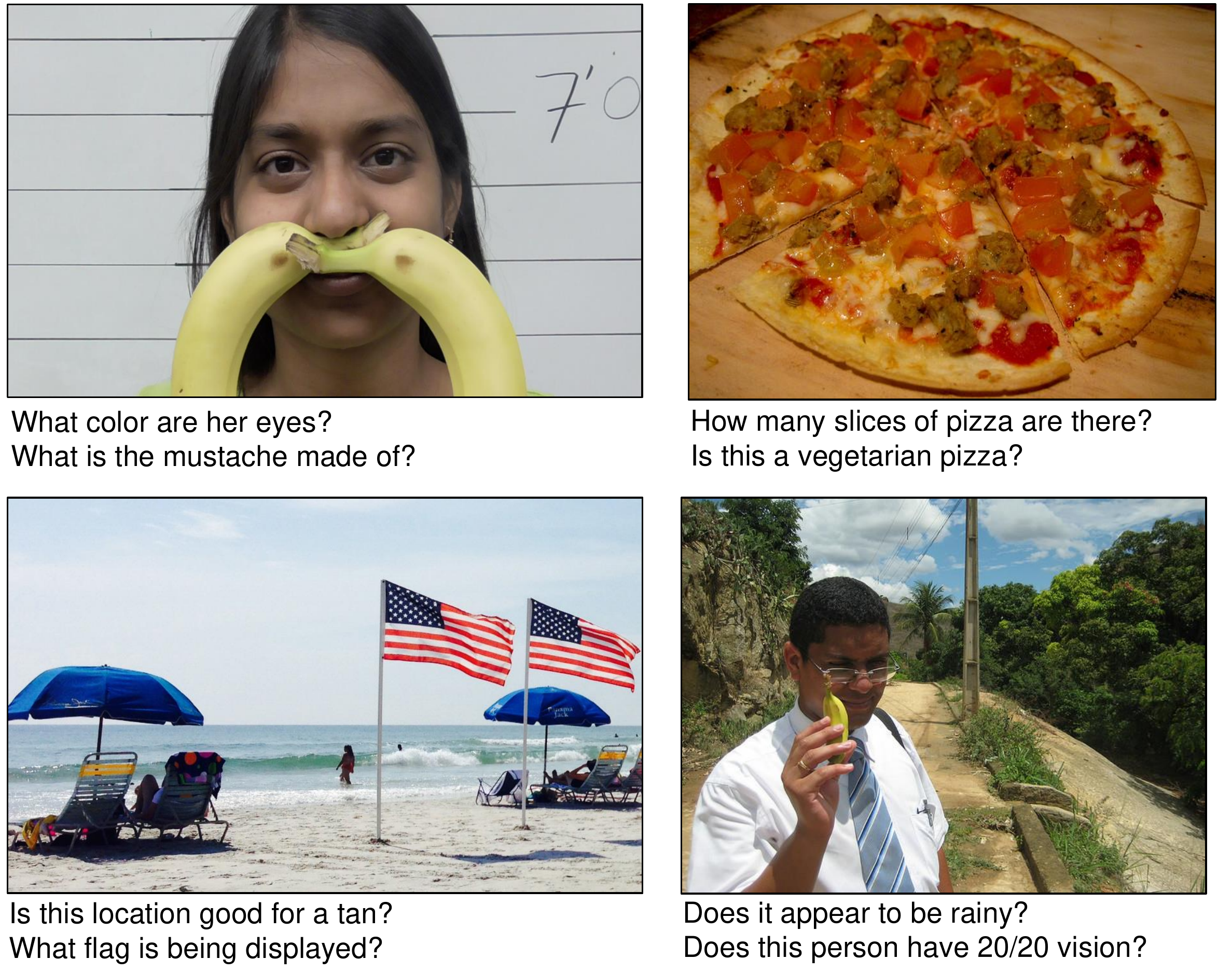}
\caption{Example images and questions in the Visual Question Answering dataset (http://visualqa.org).}
 \label{fig:teaser}
 \end{figure}
 
The task of VQA requires a machine to answer a natural language question about an image as shown in Figure \ref{fig:teaser}. Unlike the captioning task, evaluating answers to questions is relatively easy. The simplest approach is to pose the questions with multiple choice answers, much like standardized tests administered to students. Since computers don't get tired of reading through long lists of answers, we can even increase the length of the answer list. Another more challenging option is to leave the answers open-ended. Since most answers are single words such as ``yes'', ``blue'', or ``two'' evaluating their correctness is straightforward.

Is the visual question answering task challenging? The task is inherently multimodal, since it requires knowledge of language and vision. Its complexity is further increased by the fact that many questions require commonsense knowledge to answer. For instance, if you ask ``Does the man have ``20/20'' vision?'', you need the commonsense knowledge that having 20/20 vision implies you don't wear glasses. Going one step further, one might be concerned that commonsense knowledge is all that's needed to answer the questions. For example if the question was ``What color is the sheep?'', our commonsense would tell us the answer is ``white''. We may test the sufficiency of commonsense knowledge by asking subjects to answer questions without seeing the accompanying image. In this case, humans subjects did indeed perform poorly ($33\%$ correct), indicating that commonsense may be necessary but not sufficient. Similarly, we may ask subjects to answer the question given only a caption describing the image. In this case the humans performed better ($57\%$ correct), but still not as accurately as those able to view the image ($78\%$ correct). This helps indicate the VQA task requires more detailed information about an image than is typically provided in an image caption.

How do you gather diverse and interesting questions for 100,000's of images? Amazon's Mechanical Turk provides a powerful platform for crowdsourcing tasks, but the design and prompts of the experiments must be careful chosen. For instance, we ran trial experiments prompting the subjects to write questions that would be difficult for a ``toddler'', ``alien'', or ``smart robot'' to answer. Upon examination, we determined that questions written for a smart robot were most interesting given their increased diversity and difficulty. In comparison, the questions stumping a toddler were a bit too easy. We also gathered three questions per image and ensured diversity by displaying the previously written questions and stating ``Write a different question from those above that would stump a smart robot." In total over 760,000 questions were gathered \footnote{http://visualqa.org}.

\begin{figure}
\centering
\includegraphics[width=1\linewidth]{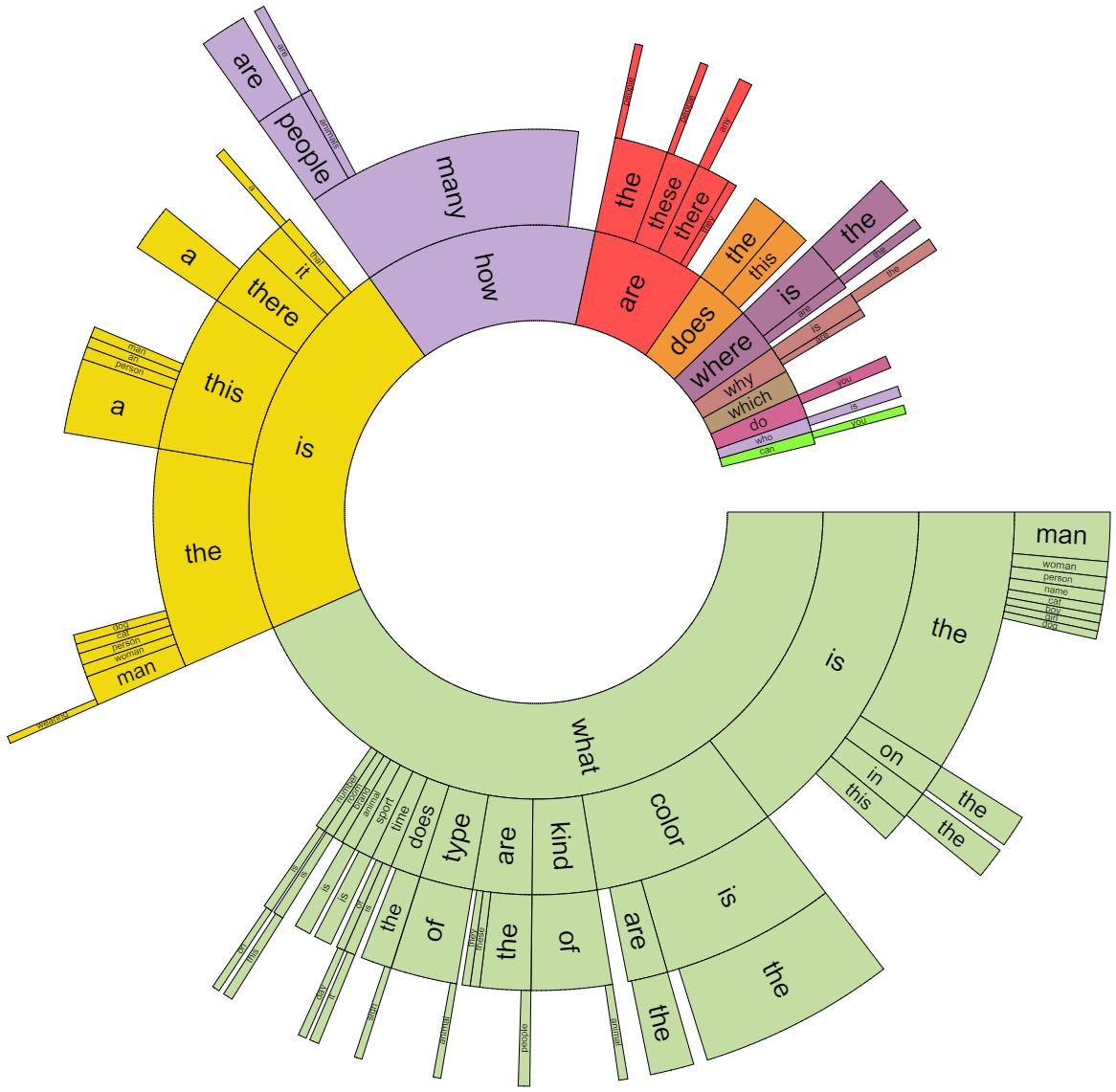}
\caption{Distribution of questions by their first four words. The ordering of the words
starts towards the center and radiates outwards. The arc length is proportional to the number of questions containing the word. White areas
indicate words with contributions too small to show.}
 \label{fig:QuestionTypes}
 \end{figure}
 
 \begin{figure*}
\centering
\includegraphics[width=1\linewidth]{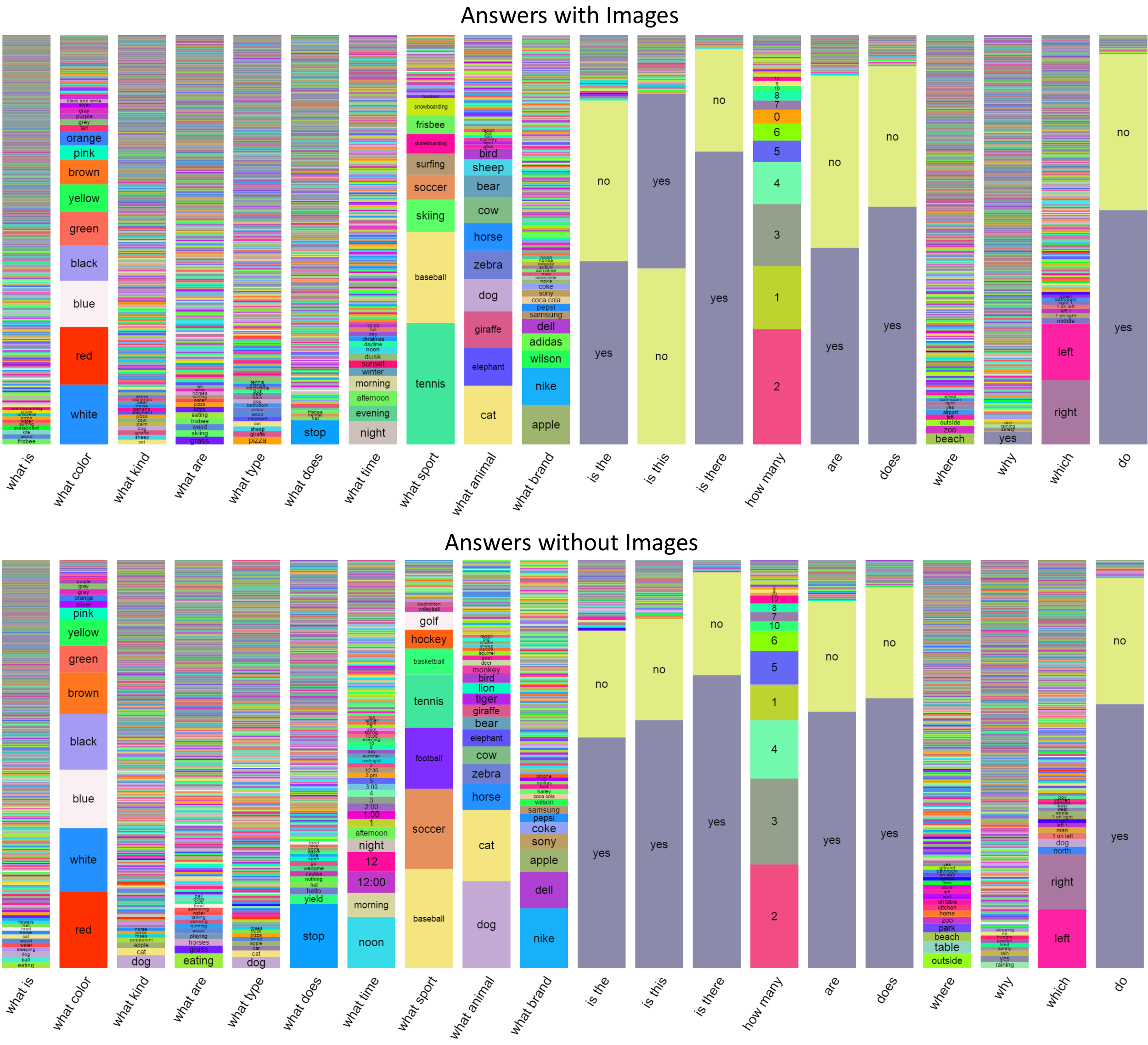}
\caption{Distribution of answers per question type when subjects provide answers when given the image (top) and when
not given the image (bottom).}
 \label{fig:answer}
 \end{figure*}
 
The diversity of questions supplied by the subjects on Amazon's Mechanical Turk is impressive. In Figure \ref{fig:QuestionTypes}, we show the distribution of words that begin the questions. The majority of questions begin with ``What'' and ``Is'', but other questions include ``How'', ``Are'', ``Does'', etc. Clearly no one type of question dominates. The answers to these questions have a varying diversity depending on the type of question. Since the answers may be ambiguous, e.g. ``What is the person looking at?" we collected ten answers per question. As shown in Figure \ref{fig:answer}, many question types are simply answered ``yes'' or ``no''. Other question types such as those that start with ``What is" have a greater variety of answers. An interesting comparison is to examine the distribution of answers when subjects were asked to answer the questions with and without looking at the image. As shown in Figure \ref{fig:answer} (bottom), there is a strong bias to many questions when subjects do not see the image. For instance ``What color'' questions invoke ``red'' as an answer, or for questions that are answered by ``yes'' or ``no'', ``yes'' is highly favored.

Finally it is important to measure the difficulty of the questions. Some questions such as ``What color is the ball?'' or ``How many people are in the room?'' may seem quite simple. In contrast, other questions such as ``Does this person expect company?'' or ``What government document is needed to partake in this activity?'' may require quite advanced reasoning to answer. Unfortunately, the difficultly of a question is in many cases ambiguous. The question's difficultly is as much dependent on the person or machine answering the question as the question itself. Each person or machine has different competencies.
 
\begin{figure*}
\centering
\includegraphics[width=1\linewidth]{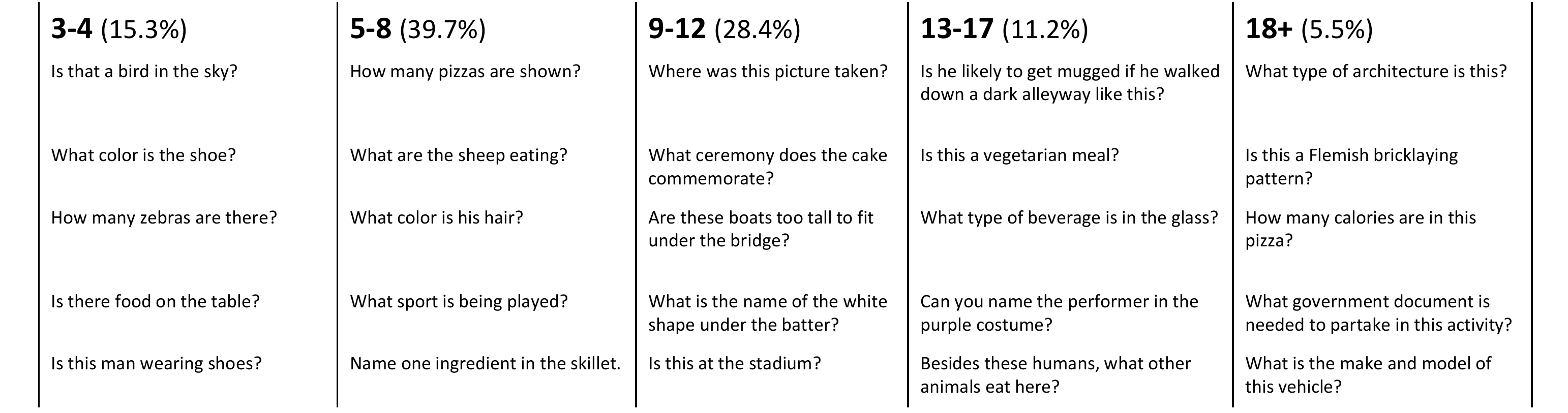}
\caption{Example questions judged to be answerable by different age groups. The percentage of questions falling into each age group is shown in parentheses.}
 \label{fig:age}
 \end{figure*}

In an attempt to gain insight into  how challenging each question is to answer, we asked human subjects to guess how old a person would need to be to answer the question. It is unlikely most human subjects have adequate knowledge of human learning development to answer the question correctly. However, this does provide an effective proxy for question difficulty. That is, questions judged to be answerable by a 3-4 year old are easier than those judged answerable by a teenager. Note, we make no claims that questions judged answerable by a 3-4 year old will actually be answered correctly by toddlers. This would require additional experiments performed by the appropriate age groups. Since the task is ambiguous, we collected ten respondences for each question. In Figure \ref{fig:age} we show several questions for which a majority of subjects picked the specified age range. Surprisingly the perceived age needed to answer the questions is fairly well distributed across the different age ranges. As expected the questions that were judged answerable by an adult (18+) generally need specialized knowledge, where those answerable by a toddler (3-4) are more generic.

\section{Abstract Scenes}

The visual question answering task requires a variety of skills. The machine must be able to understand the image, interpret the question and reason about the answer. For many researchers exploring AI, they may not be interested in exploring the low-level tasks involved with perception and computer vision. Many of the questions may even be impossible to solve given the current capabilities of state-of-the-art computer vision algorithms. For instance the question ``How many cellphones are in the image?'' may not be answerable if the computer vision algorithms cannot accurately detect cellphones. In fact, even for state-of-the-art algorithms many objects are difficult to detect, especially small objects \cite{coco}.

\begin{figure}[h]
\centering
\includegraphics[width=1\linewidth]{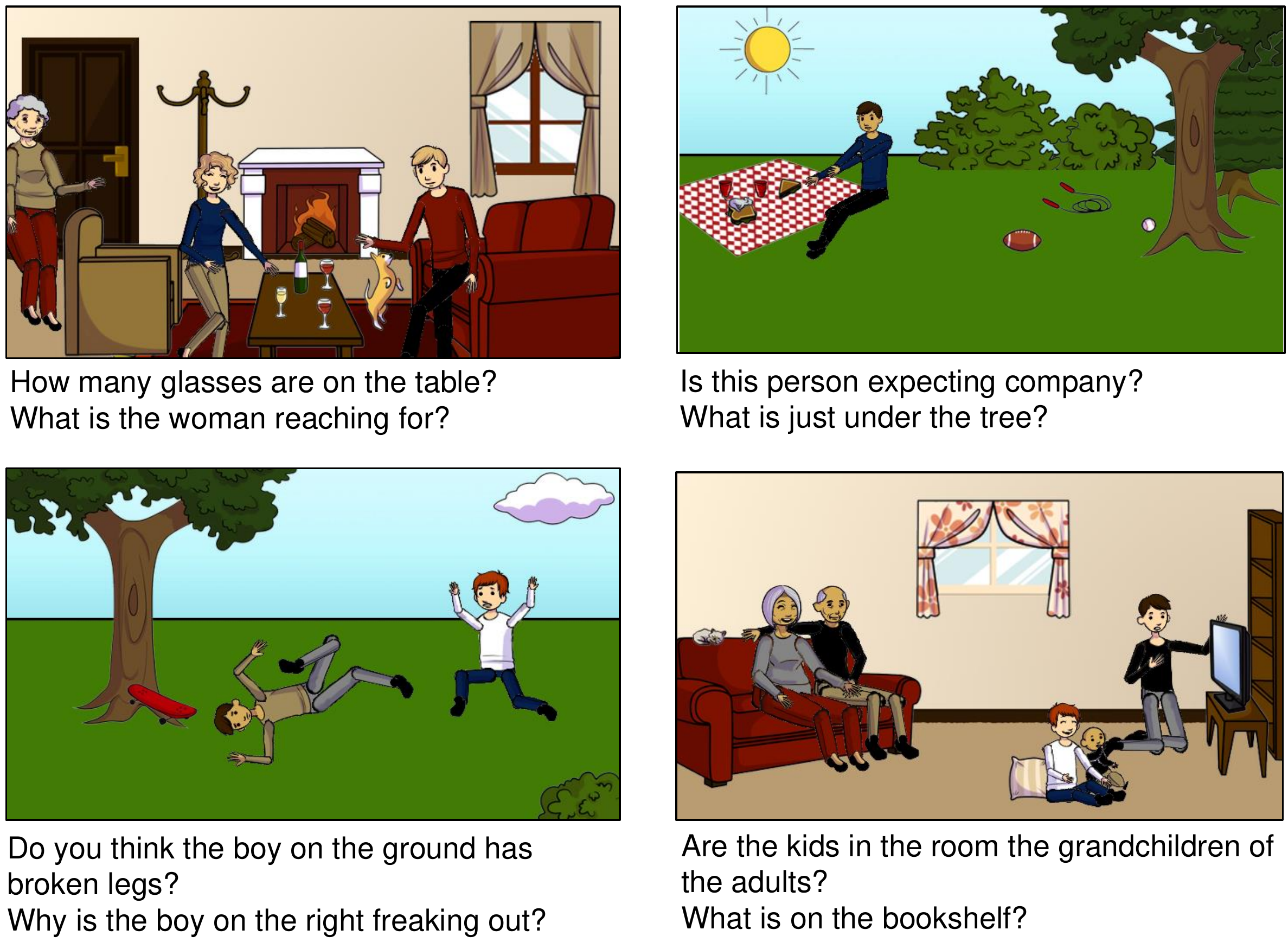}
\caption{Example abstract scenes and their questions in the Visual Question Answering dataset (http://visualqa.org).}
 \label{fig:abstract}
 \end{figure}

To enable multiple avenues for researching VQA, we introduce abstract scenes into the dataset \cite{Antol2014,ZitnickCVPR2013,ZitnickICCV2013,VedantamPAMI2015}. Abstract scenes or cartoon images are created from sets of clip art, Figure \ref{fig:abstract}. The scenes are created by human subjects using a graphical user interface that allows them to arrange a wide variety objects. For clip art depicting humans, their poses and expression may also be changed. Using the interface a wide variety of scenes can be created including ordinary scenes, scary scenes, or funny scenes. Since the type of clip art and it's properties are exactly known, the problem of recognizing objects and their attributes is greatly simplified. This provides researchers an opportunity to more directly study the problems of question understanding and answering. Once computer vision algorithms ``catch up'', perhaps some of the techniques developed for abstract scenes can be applied to real images. The abstract scenes may be useful for a variety of other tasks as well, such as learning common sense knowledge \cite{ZitnickICCV2013,Antol2014,NEIL,divvala2014learning,Vedantam2015}.

\section{Discussion}

While visual question answering appears to be a promising approach to measuring machine intelligence for multimodal tasks, it may prove to have unforseen shortcomings. We've explored several baseline algorithms that perform poorly when compared to human performance. As the dataset is explored, it is possible that solutions may be found that don't require ``true AI''. However, using proper analysis we hope to continuously update the dataset to reflect the current progress of the field. As certain question or image types become too easy to answer we can add new questions and images. Other modalities may also be explored such as audio and text-based stories \cite{zettlemoyer_kdd14,zettleymoyer_acl13,weston_qa,richardson2013mctest}.

In conclusion, we believe designing a multimodal challenge is essential for accelerating and measuring the progress of AI. Visual question answering offers one approach for designing such challenges that allows for easy evaluation while maintaining the difficultly of the task. As the field progresses our tasks and challenges should be continuously reevaluated to ensure they are of appropriate difficultly given the state of research. Importantly, these tasks should be designed to push the frontiers of AI research, and help ensure their solutions lead us towards systems that are truly ``AI complete''.

{\small
\bibliographystyle{ieee}
\bibliography{vqa_aimag}
}

\end{document}